\documentclass{article}

\PassOptionsToPackage{numbers,sort}{natbib}

\usepackage[preprint]{neurips_2026}

\usepackage[utf8]{inputenc} 
\usepackage[T1]{fontenc}    
\usepackage{hyperref}       
\usepackage{url}            
\usepackage{booktabs}       
\usepackage{amsfonts}       
\usepackage{nicefrac}       
\usepackage{microtype}      
\usepackage{xcolor}         

\usepackage{mathtools}

\title{Beyond Autoregressive RTG: Conditioning via Injection Outside Sequential Modeling in Decision Transformer}

\author{%
  Yongyi~Wang \\
  School of Computer Science\\
  Peking University\\
  Beijing, China 100871 \\
  \texttt{wangyongyi@pku.edu.cn} \\
  \And
  Hanyu~Liu \\
  School of Computer Science\\
  Peking University\\
  Beijing, China 10871\\
  \AND
  Lingfeng~Li \\
  School of Computer Science\\
  Peking University\\
  Beijing, China 10871
  \And
  Bozhou~Chen \\
  School of Computer Science\\
  Peking University\\
  Beijing, China 10871\\
  \And
  Ang~Li \\
  School of Computer Science\\
  Peking University\\
  Beijing, China 10871\\
  \And
  Qirui~Zheng \\
  School of Computer Science\\
  Peking University\\
  Beijing, China 10871\\
  \And
  Xionghui~Yang \\
  School of Computer Science\\
  Peking University\\
  Beijing, China 10871\\
  \And
  Chucai~Wang \\
  School of Computer Science\\
  Peking University\\
  Beijing, China 10871\\
  \And
  Wenxin~Li \\
  School of Computer Science\\
  Peking University\\
  Beijing, China 10871\\
}

\begin{document}

\maketitle

\begin{abstract}
Decision Transformer (DT) formulates offline reinforcement learning as autoregressive sequence modeling, achieving promising results by predicting actions from a sequence of Return-to-Go (RTG), state, and action tokens. However, RTG is a scalar that summarizes future rewards, containing far less information than typical state or action vectors, yet it consumes the same computational budget per token. Worse, the self-attention cost of Transformers grows quadratically with sequence length, so including RTG as a separate token adds unnecessary overhead.
We propose SlimDT, which removes RTG from the autoregressive sequence. Instead, we inject RTG information into the state representations before the sequential modeling step, allowing the Transformer to process only a compact (state, action) sequence. This reduces the sequence length by one-third, directly improving inference efficiency. On the D4RL benchmark, SlimDT surpasses standard DT across various tasks and achieves performance comparable to existing state-of-the-art methods. Decoupling a sparse conditioning signal from an information-rich sequence thus yields both computational gains and higher task performance. 
\end{abstract}

\section{Introduction}
Offline Reinforcement Learning (RL) has emerged as a critical paradigm for learning policies from static, pre-collected datasets, bypassing the need for costly or unsafe online interaction. This setting is particularly valuable in real-world domains such as industrial control and robotic operation, where data collection is expensive and safety constraints are stringent. Among recent advances, the Decision Transformer (DT) \cite{chen2021decision} has introduced a radically different perspective: it casts offline RL as an autoregressive sequence modeling problem. Concretely, DT uses a Transformer \cite{vaswani2017attention} decoder to process a triplet sequence of Return-to-Go (RTG), state, and action tokens at each time step, and predicts the next action conditioned on the past trajectory.

The DT framework has demonstrated promising results across various benchmarks. Its key component is the RTG, a scalar that specifies the desired cumulative reward from the current step onward. During training, RTG helps the model distinguish possible high-quality trajectories; during inference, it serves as an adjustable target that guides action generation. In the standard DT design, the RTG is treated on par with the state and action vectors: it occupies one position in the input sequence and undergoes the same self-attention computation as every other token.

However, this design choice raises an important question: Does a low-information scalar deserve the same computational budget as rich state and action representations? Unlike state vectors (which capture detailed environment observations) or action vectors (which encode control decisions), RTG is merely a single number summarizing future rewards: its information content is drastically lower. Moreover, the Transformer’s inference cost grows quadratically with sequence length. Including RTG as a separate token increases the length by 50\% (from two to three tokens per step), incurring non-negligible computational overhead while providing minimal information gain.

In this paper, we challenge the autoregressive treatment of RTG. We argue that conditioning information should not be forced into the same sequential pipeline as the primary dynamics data. Instead, we propose SlimDT, a simple but effective modification to the DT architecture. SlimDT removes the RTG token from the autoregressive sequence altogether. Before or after sequential modeling, we inject (e.g., via cross-attention) RTG information into the encoded state representations, so that the Transformer only processes a compact sequence of (state, action) pairs. This reduces the effective sequence length by $1/3$, directly lowering quadratic inference cost while improving task performance.

We evaluate SlimDT on challenging D4RL benchmarks \cite{fu2020d4rl}: the Adroit suite (pen, door, hammer, relocate) for dexterous manipulation, and the MuJoCo suite (hopper, halfcheetah, walker) for continuous control. Experimental results show that our method outperforms the standard DT and achieves performance comparable to state-of-the-art methods across all tasks, while providing consistent gains in inference efficiency. These findings suggest that decoupling a low-information conditioning signal from the autoregressive sequence is not only computationally rational but also beneficial for final performance.

Our main contributions are summarized as follows:
\begin{itemize}
\item We identify a previously overlooked inefficiency in the DT: treating the low-information RTG scalar as an equal token in the sequence leads to unnecessary computational overhead.
\item We propose SlimDT, a conditioning paradigm that injects RTG information into state representations, eliminating the RTG token entirely. This reduces the autoregressive sequence length by $1/3$, directly lowering quadratic inference cost while improving performance.
\item We demonstrate on D4RL that SlimDT improves upon standard DT performance, and is even comparable to state-of-the-art methods.
\end{itemize}
The rest of the paper is organized as follows: Section \ref{sec:relwork} reviews related works, mainly offline RL algorithms; Section \ref{sec:prelim} introduces relevant background knowledge about DT and the conditional injection module used in our design; Section \ref{sec:framework} elaborates on the proposed SlimDT architecture, including two variants (pre-conditioning and post-conditioning) and different choices for the condition injector; Section \ref{sec:exp} presents the main results of SlimDT on D4RL and the corresponding ablation experiments, along with key findings; Section \ref{sec:concl} summarizes the contributions, limitations, and potential future directions.

\section{Related Works}
\label{sec:relwork}
\subsection{Offline Reinforcement Learning via Dynamic Programming}
Offline RL learns policies purely from pre-collected static datasets without online environment interaction. Its key challenge is policy-induced distribution shift: learned policies may produce out-of-distribution (OOD) state-action pairs absent from the behavioral dataset, causing severe value extrapolation errors.
Existing offline RL via dynamic programming (DP) approaches can be categorized into four mainstream paradigms. 

\emph{Policy constraint methods}: including BCQ \cite{fujimoto2019off}, BEAR \cite{kumar2019stabilizing}, and TD3+BC \cite{fujimoto2021minimalist}, restrict policies to the support of offline behavioral data and prevent risky out-of-distribution (OOD) exploration.

\emph{Conservative value regularization methods}: (CQL \cite{kumar2020conservative}, EDAC \cite{an2021uncertainty}) alleviate OOD value overestimation by imposing conservative penalties on unseen state-action pairs.

\emph{Constraint-free implicit value learning}: extracts optimal behaviors via regression-based value estimation without explicit policy optimization. As a representative method, IQL \cite{kostrikov2021offline} adopts expectile regression for advantage fitting to avoid OOD value evaluation. Different from IQL, XQL \cite{garg2023extreme} performs extreme value Q-learning through reweighted value iteration, asymptotically sharpening value estimates to suppress suboptimal OOD behaviors.

\emph{Model-based offline RL methods}: such as MOPO \cite{yu2020mopo}, MOReL \cite{kidambi2020morel}, and COMBO \cite{yu2021combo}, train uncertainty-aware dynamics models to generate valid synthetic trajectories, remedying the limited state-action coverage of offline datasets.
Despite these advances, most value-based and model-based offline RL methods suffer from bootstrapping error accumulation in iterative dynamic programming updates and exhibit high sensitivity to hyperparameters.

\subsection{Offline Reinforcement Learning via Sequence Modeling}
Departing from offline RL algorithms based on DP, the DT \cite{chen2021decision} frames RL as sequence modeling.
By leveraging a Transformer \cite{vaswani2017attention} to autoregressively predict actions conditioned on past observations, actions, and RTGs, DT avoids bootstrapping and the Markov assumption, establishing a robust, trajectory modeling paradigm.
GDT \cite{furutageneralized} later provided a unifying theoretical framework of Hindsight Information Matching (HIM).
Subsequent research has targeted specific limitations \cite{wang2026decoupling} of DT. 

\emph{Suboptimal Trajectory Stitching}: Enhanced by generating waypoints (WT \cite{badrinath2023waypoint}), adaptive context lengths (EDT \cite{wu2023elastic}), Q-learning relabeling (QDT \cite{yamagata2023q}), hierarchical joint optimization (ADT \cite{ma2024rethinking}). These methods enable DT to combine fragments of suboptimal trajectories into near-optimal policies, overcoming the stitching difficulty inherent in return-conditioned sequence models.

\emph{Stochasticity \& Over-optimism}: Addressed by clustering returns (ESPER \cite{paster2022you}) or disentangling latent variables (DoC \cite{yangdichotomy}), advantage conditioning (ACT \cite{gao2024act}). Such techniques reduce the harmful effects of environment randomness and inflated value estimates, leading to more robust policy learning.

\emph{Online Training \& Adaptation}: Explored via online fine-tuning (ODT \cite{zheng2022online}), auxiliary critics (CGDT \cite{wang2024critic}), RL gradient infusion (TD3+ODT \cite{yan2024reinforcement}, GRPO-DT \cite{luo2026online}), value function guidance (VDT \cite{zheng2025value}). These approaches improve DT’s out-of-distribution generalization and sample efficiency when transitioning from offline pre-training to online interaction.

Models other than Transformer have also been explored, including convolutional models (DC \cite{kim2023decision}), Mamba \cite{gu2024mamba} (Decision Mamba \cite{lv2024decision,huang2024decision}), and hybrid designs (LSDT \cite{wang2025long}, DHi \cite{huangless}). These alternatives aim to reduce the quadratic complexity of self-attention or enhance local dependency modeling while maintaining competitive performance.
\section{Preliminaries}
\label{sec:prelim}
\subsection{Decision Transformer}
DT is an offline reinforcement learning method. Unlike classic DP-based Offline RL approaches, it leverages reward information in trajectories by taking the sum of these future rewards, namely the RTG, as the condition for generating actions to guide decision-making. Therefore, it is also a Return-Conditioned Supervised Learning (RCSL) method based on sequence modeling.

For each timestep $t$, a trajectory slice $\tau_{t-k+1:t}$ of context length $k$ has the form:
\begin{equation}
    \tau_{t-k+1:t}\coloneqq(R_{t-k+1},s_{t-k+1:t},a_{t-k+1:t},\dots,R_{t},s_t)
\end{equation}
where the RTG $R_t\coloneqq\sum_{i=t}^T r_i$.

During training, an offline dataset containing complete trajectories is used. The RTG $R_t$ at each time step is preprocessed. Then $R_t, s_t, a_t$ are respectively encoded into an alternating sequence of $(R^\prime_t,s^\prime_t,a^\prime_t)\coloneqq (E_R(R_t), E_s(s_t), E_a(a_t))$ (tokens of the same dimension) via different linear layers $E_R, E_s, E_a$. Then they are fed into the GPT2 architecture \cite{radford2018improving,radford2019language} to predict the actions $\hat{a}_{t-k+1:t}$.
\begin{equation}
\hat{a}_{t-k+1:t}=GPT(R_{t-k+1}^\prime,s_{t-k+1}^\prime,a_{t-k+1}^\prime,\dots,R_t^\prime,s_t^\prime)
\end{equation}
For continuous control, MSE is used between predicted and ground-truth actions, while cross-entropy applies to discrete cases.
During inference, since the trajectory has not been fully generated, an expected RTG $\hat{R}_0$ needs to be selected according to the characteristics of the problem at the initial stage. As the inference proceeds, the single-step reward is sequentially subtracted to estimate the subsequent RTG values. The initial RTG is generally set to a value higher than the maximum theoretical return, so as to prompt the Transformer to tend to generate trajectories with higher actual rewards. The input during the inference phase is in the form of
\begin{equation}
    \hat{\tau}_{t-k+1:t}\coloneqq(\hat{R}_{t-k+1},s_{t-k+1:t},a_{t-k+1:t},\dots,\hat{R}_{t},s_t)
\end{equation}
where $\hat{R}_t=\hat{R}_0-\sum_{i=0}^{t-1}r_i$.
Generally, either excessively high or low RTG impairs the performance of DT when tested in simulation or real-world environments.

\subsection{Cross-Attention}
Cross-attention is a fundamental self-attention variant for cross-sequence information fusion and asymmetric feature alignment \cite{vaswani2017attention}. Unlike self-attention that adopts $Q, K, V$ from a single sequence for intra-sequence modeling, cross-attention decouples feature sources: queries ($\textcolor{blue}{Q}$) come from the \textcolor{blue}{target} sequence to be enhanced, while keys ($\textcolor{orange}{K}$) and values ($\textcolor{orange}{V}$) are from the guiding \textcolor{orange}{source} sequence. Standard cross-attention take the form of:
\begin{equation}
\label{eq:crossattn}
\text{CrossAttn}(\textcolor{blue}{Q},\textcolor{orange}{K},\textcolor{orange}{V})\coloneqq\text{softmax}\left(\frac{\textcolor{blue}{Q}\textcolor{orange}{K^\top}}{\sqrt{d}}\right)\textcolor{orange}{V}
\end{equation}
It first computes the similarity between \textcolor{blue}{target $Q$} and \textcolor{orange}{source $K$}, and normalizes the scores via softmax to generate cross-attention weights, which reflect the cross-sequence correspondence. These weights aggregate \textcolor{orange}{source $V$} features and inject source contextual information into target representations. This mechanism enables conditional feature modulation, serving as a core component for sequence translation, cross-modal fusion and conditional generation tasks.

\subsection{Adaptive Layer Normalization}
\label{subsec:adaln}
Layer Normalization (LN) \cite{ba2016layer} was introduced to stabilize training in RNNs by reducing internal covariate shift, proving effective for sequential and Transformer-based architectures. Building on LN and conditional normalization from earlier generative models \cite{de2017modulating,karras2019style,dhariwal2021diffusion}, DiT \cite{peebles2023scalable} popularized the modern adaLN module, where a shared MLP maps conditioning vectors (e.g., timestep or class embeddings) to layer-specific modulation parameters (scale $\gamma$ and bias $\beta$). This mechanism dynamically injects conditional information by replacing static affine transformations of LN with adaptive, condition-dependent scaling and shifting.
\begin{equation}
\label{eq:adaln}
\begin{aligned}
    \gamma(z), \beta(z)&=\text{MLP}(z) \\
    \text{adaLN}(x,z)&=\gamma(z)\odot\frac{x-\mu(x)}{\sqrt{\sigma^2(x)+\epsilon}}+\beta(z)
\end{aligned}
\end{equation}
where $\mu$ is the mean value of each dimension of $x$, $\sigma^2$ is the variance, and $\odot$ denote the Hadamard product. Generally, a single-layer linear transformation can be adopted for MLP.
\section{Proposed Framework}
\label{sec:framework}
\subsection{SlimDT Design}
\begin{figure}[t] 
    \centering
    \includegraphics[width=1.0\linewidth]{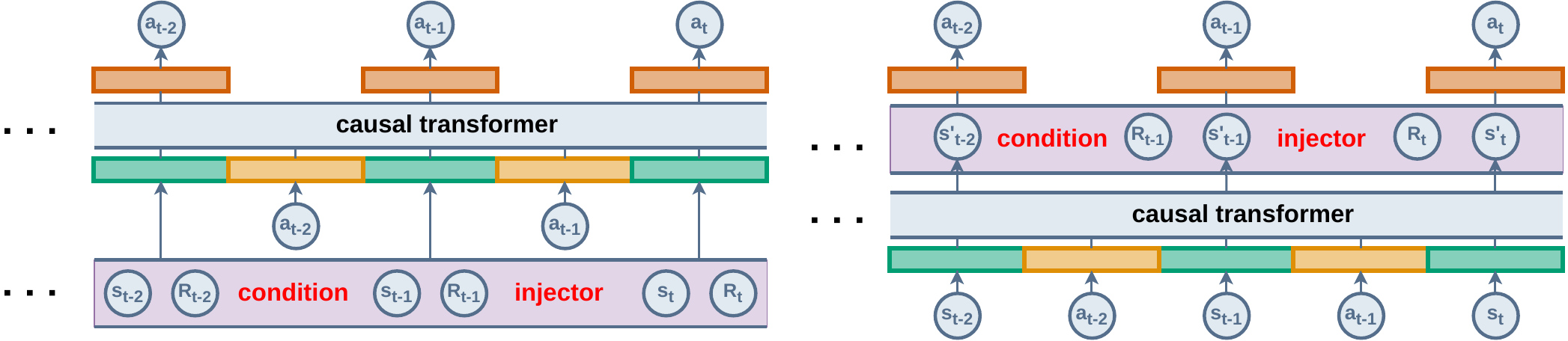}
    \caption{SlimDT architecture. (Left) Pre-conditioning: the RTG sequence $R_{t-k+1:t}$ is fused with the state sequence $s_{t-k+1:t}$ via the condition injector, then fed into GPT together with the action sequence $a_{t-k+1:t-1}$, which autoregressively outputs the action predictions $\hat{a}_{t-k+1:t}$. (Right) Post-conditioning: the state sequence $s_{t-k+1:t}$ and the action sequence are first fed into GPT to generate the encoded trajectory $s'_{t-k+1:t}$, which is then fused with the RTG sequence $R_{t-k+1:t}$ through the condition injector to generate action predictions.}
    \label{fig:slimdt}
\end{figure}

Given that DT treats the sparse-information RTG tokens on par with the information-rich state and action embeddings, consuming equal computational resources and suffering from inefficiency due to increased sequence length, we propose SlimDT, streamlined architectures of its two variants: pre-cond and post-cond are illustrated in Fig.~\ref{fig:slimdt}.

Specifically, in SlimDT-pre-cond, the RTG sequence $R_{t-k+1:t}$ is first fused with the state sequence $s_{t-k+1:t}$ via a condition injector, producing a return-conditioned state representation $s^\prime_{t-k+1:t}$. This representation is then interleaved with the action sequence and fed into the Transformer for action prediction.
\begin{equation}
s^\prime_{t-k+1:t}=\text{CondInject}(R_{t-k+1:t},s_{t-k+1:t})
\end{equation}
\begin{equation}
\hat{a}_{t-k+1:t}=\text{GPT}(s_{t-k+1}^\prime,a_{t-k+1}^\prime,\dots,s_{t-1}^\prime,a_{t-1}^\prime,s_t^\prime)
\end{equation}

In SlimDT-post-cond, the interleaved state-action sequence is first fed into the Transformer to produce encoded trajectory tokens $s^\prime_{t-k+1:t}$, which are then modulated by the RTG sequence to generate action predictions.
\begin{equation}
s^\prime_{t-k+1}=\text{GPT}(s_{t-k+1},a_{t-k+1},\dots,s_{t-1},a_{t-1},s_t)
\end{equation}
\begin{equation}
\hat{a}_{t-k+1:t}=\text{CondInject}(R_{t-k+1:t},s^\prime_{t-k+1:t})
\end{equation}

Except for the different conditional injection methods (CondInject module for SlimDT and Transformer for DT), the training and inference processes of SlimDT and DT are exactly the same.
SlimDT reduces the sequence length input to GPT to $2/3$ of the original. Considering the quadratic time complexity of GPT, with $k$ kept constant and ignoring the overhead of the condition injector, the theoretical inference time consumption of SlimDT becomes $4/9$ that of DT.

For the convenience of discussion, unless otherwise specified, SlimDT appearing hereinafter all refers to SlimDT-pre-cond.

\subsection{Different Options for RTG Condition Injector}
Within the time window $t-k+1:t$, the information provided by the RTG sequence $R_{t-k+1:t}$ for decision-making can be divided into two parts: the expected future remaining return $R_t$ at time step $t$, and the rewards $r_{t-k+1:t-1}$ generated at each time step. Among them, $R_t$ mainly serves to guide the generation of high-value actions, while $r_{t-k+1:t-1}$ can supplement some missing information in the state representation for decision-making in certain scenarios. For instance, sudden changes in rewards at certain moments indicate the occurrence of important events that are meaningful for decision-making, and such events may lack information representation implied in the state sequence.

Condition injectors fall into two categories: point-to-point and sequence-to-sequence. The former only injects the RTG at the corresponding time step into the state, such as the concatenation and adaLN method described later, and generally achieves higher computational efficiency. The latter modulates the state by leveraging RTG information across the entire time window, like the cross-attention approach introduced below; it is generally less computationally efficient than the former but enables more thorough information fusion, and it is also more conducive to capturing event information useful for decision-making contained in the reward signal sequence.

Next, we introduce three typical modules that can implement conditional information injection: concatenation, adaLN, and cross-attention respectively. The first two belong to point-to-point, and the latter belongs to sequence-to-sequence.

\subsubsection{Concatenation}
The simplest way to inject the reward condition $R_t$ into $s_t$ is to directly concatenate the vectors.
However, in practice, this method yields unsatisfactory results. Therefore, we first apply a linear mapping $E_R$ on $R_t$ to a higher dimension, then concatenate it with $s_t$, and finally use another linear encoder $E_{s^\prime}$ to produce the input for the Transformer:
\begin{equation}
s^\prime_t=E_{s^\prime}(\text{Concat}(E_R(\textcolor{orange}{R_t}),\textcolor{blue}{s_t}))
\end{equation}
\subsubsection{Adaptive Layer Normalization}
As introduced in Sec. \ref{subsec:adaln}, adaLN is a module originally designed for conditional injection. It can also be used to inject RTG into the state representation as a modulation signal.
\begin{equation}
s^\prime_t=\text{adaLN}(E_s(\textcolor{blue}{s_t}),E_R(\textcolor{orange}{R_t}))
\end{equation}
\subsubsection{Cross Attention}
\label{subsubsec:crossattn}
The standard cross-attention mechanism given by Eq. \ref{eq:crossattn} takes the sequences corresponding to $K$ and $V$ as the condition for injecting into the sequence corresponding to $Q$. When migrated to the scenario of SlimDT, $K$ and $V$ correspond to the RTG sequence, while $Q$ corresponds to the state sequence.

However, we find empirically that SlimDT performs poorly when using standard cross-attention as the conditioning mechanism. We attribute this to the fact that the RTG sequence, which serves as the conditioning signal, contains substantially less information than the state sequence. In standard cross-attention, the information-poor RTG sequence occupies both the $K$ and $V$ matrices, while the information-rich state sequence is only assigned to $Q$. This misallocation undermines the representational capacity of the network.

By the original design of cross-attention, $Q$ corresponds to the decoder sequence to be modulated (which is typically information-sparser), whereas $K$ and $V$ come from the encoder sequence that provides richer contextual information for modulation. However, the situation in DT is exactly the opposite: the \textcolor{orange}{conditioning RTG sequence $R_{t-k+1:t}$} is information-sparse, while the \textcolor{blue}{state sequence $s_{t-k+1:t}$ to be modulated} is information-rich. This contradiction explains why standard cross attention performs poorly when used as a condition injector in SlimDT.

Given the specific conditioning requirements of SlimDT, we propose two variants of cross‑attention:
\begin{equation}
s^\prime_{t-k+1:t}=\text{CrossAttn}_{\textcolor{orange}{Q}\rightarrow\textcolor{blue}{KV}}(\textcolor{orange}{Q},\textcolor{blue}{K},\textcolor{blue}{V})\coloneqq\text{softmax}\left(\frac{\textcolor{orange}{Q}\textcolor{blue}{K^\top}}{\sqrt{d}}\right)\textcolor{blue}{V}
\end{equation}
where $\textcolor{orange}{Q}=\mathbf{Q}E_R(\textcolor{orange}{R_{t-k+1:t}})$, $\textcolor{blue}{K}=\mathbf{K}E_s(\textcolor{blue}{s_{t-k+1:t}})$, $\textcolor{blue}{V}=\mathbf{V}E_s(\textcolor{blue}{s_{t-k+1:t}})$, and
\begin{equation}
s^\prime_{t-k+1:t}=\text{CrossAttn}_{\textcolor{orange}{K}\rightarrow\textcolor{blue}{QV}}(\textcolor{blue}{Q},\textcolor{orange}{K},\textcolor{blue}{V})\coloneqq\text{softmax}\left(\frac{\textcolor{blue}{Q}\textcolor{orange}{K^\top}}{\sqrt{d}}\right)\textcolor{blue}{V}
\end{equation}
where $\textcolor{blue}{Q}=\mathbf{Q}E_s(\textcolor{blue}{s_{t-k+1:t}})$, $\textcolor{orange}{K}=\mathbf{K}E_R(\textcolor{orange}{R_{t-k+1:t}})$, $\textcolor{blue}{V}=\mathbf{V}E_s(\textcolor{blue}{s_{t-k+1:t}})$.

In these two variants of cross attention, one of $Q$ and $K$ corresponds to the RTG sequence, while the other, together with $V$, corresponds to the state sequence. This is equivalent to first calculating the correlation between each pair of elements in the RTG sequence and the state sequence, then using these correlation scores to modulate the state sequence so as to inject return condition information.
\section{Experiments}
\label{sec:exp}
\subsection{Dataset, Baseline and Implementation}
\textbf{Dataset}: We evaluate SlimDT on the D4RL benchmark \cite{fu2020d4rl}, covering both the Adroit suite and MuJoCo locomotion tasks. All scores are normalized via the official D4RL evaluation interface.

Adroit includes four dexterous manipulation tasks (Pen, Hammer, Door, Relocate) with the Shadow Hand, where we use two standard offline variants per task: Human (limited teleoperated demonstrations) and Cloned (1:1 mixture of human and behavior-cloned rollouts). The sparse rewards, high action dimensionality, and limited coverage make Adroit a challenging testbed. 

For locomotion, we evaluate on Hopper, Walker2d, and HalfCheetah, using three dataset types per environment: medium (generated by a medium-performance policy), medium-replay (replay buffer of a medium-trained agent), and medium-expert (1:1 mixture of medium and expert trajectories). 

\textbf{Baseline}: We compare our approach with existing state-of-the-art offline RL approaches, each excelling in
specific domain tasks. For value-based approaches, we include CQL \cite{kumar2020conservative}, IQL \cite{kostrikov2021offline}, and BCQ \cite{fujimoto2019off}.
For supervised learning approaches, we include DT \cite{chen2021decision}, StAR \cite{shang2022starformer}, GDT \cite{furutageneralized}, DC \cite{kim2023decision}, DHi \cite{huangless}. 

\textbf{Implementation}: Our DT and SlimDT are both based on d3rlpy \cite{seno2022d3rlpy}, an elegant open-source offline RL library with comprehensive documentation and a well-designed interface.

Experiments show that for relatively simple MuJoCo series tasks, combining SlimDT-post-cond with adaLN as condition injector achieves the optimal performance. For more challenging Adroit tasks, the optimal results are generally obtained by adopting SlimDT-pre-cond in combination with cross-attention as condition injector.

\subsection{Main Results}
\begin{table}[h]
    \centering
    \begin{tabular}{lrrrrrrrrrr}
        \toprule
        \textbf{MuJoCo} & \textbf{CQL} & \textbf{IQL} & \textbf{BCQ} & \textbf{BC} & \textbf{DT} & \textbf{StAR} & \textbf{GDT}  & \textbf{DC} & \textbf{DHi} & \textbf{Ours}\\
        \midrule
        Hc-me & 91.6 & 86.7 & 69.6 & 55.2 & 86.8 & 93.7 & 93.2 & 93.0 & \textbf{94.2} & \textbf{94.2}$\pm$0.2 \\
        Hp-me & 105.4 & 91.5 & 109.1 & 52.5 & 107.6 & 111.1 & 111.1 & 110.4 & \textbf{111.7} & 111.1$\pm$0.2 \\
        Wk-me & 108.8 & 109.6 & 67.3 & 107.5 & 108.1 & 109.0 & 107.7 & \textbf{109.6} & \textbf{109.6} & 109.5$\pm$0.5 \\
        Hc-m & \textbf{49.2} & 47.4 & 41.5 & 42.6 & 42.6 & 42.9 & 42.9 & 43.0 & 43.4 & 43.0$\pm$0.4 \\
        Hp-m & 69.4 & 66.3 & 65.1 & 52.9 & 67.6 & 59.5 & 77.6 & 92.5 & 90.1 & \textbf{99.4}$\pm$0.6 \\
        Wk-m & \textbf{83.0} & 78.3 & 52.0 & 75.3 & 74.0 & 73.8 & 76.5 & 79.2 & 79.9 & 78.6$\pm$0.3 \\
        Hc-mr & \textbf{45.5} & 44.2 & 34.8 & 36.6 & 36.6 & 36.8 & 40.5 & 41.3 & 41.5 & 37.8$\pm$0.5 \\
        Hp-mr & 95.0 & 94.7 & 31.1 & 18.1 & 82.7 & 29.2 & 85.3 & 94.2 & \textbf{97.7} & 92.5$\pm$0.6 \\
        Wk-mr & 77.2 & 73.9 & 13.7 & 32.3 & 79.4 & 39.8 & 77.5 & 76.6 & \textbf{81.2} & 77.6$\pm$1.1 \\
        \midrule
        Avg. & 80.6 & 77.0 & 53.8 & 52.6 & 76.2 & 66.2 & 79.1 & 82.2 & \textbf{83.3} & 82.6 \\
        \toprule
        \textbf{Adroit} & \textbf{CQL} & \textbf{IQL} & \textbf{BCQ} & \textbf{BC} & \textbf{DT} & \textbf{StAR} & \textbf{GDT}  & \textbf{DC} & \textbf{DHi} & \textbf{Ours} \\
        \midrule
        P-h & 37.5 & 71.5 & 66.9 & 63.9 & 79.5 & 77.9 & 92.5 & 93.8 & 86.6 & \textbf{94.4}$\pm$8.7 \\
        H-h & 4.4 & 1.4 & 0.9 & 1.2 & 3.7 & 3.7 & 5.5 & \textbf{43.0} & 31.2 & 39.1$\pm$3.9 \\
        D-h & 9.9 & 4.3 & -0.1 & 2.0 & 14.8 & 1.5 & 20.6 & 22.6 & \textbf{25.2} & 21.3$\pm$1.3 \\
        P-c & 39.2 & 37.3 & 50.9 & 37.0 & 75.8 & 33.1 & 86.2 & 98.2 & 89.1 & \textbf{99.3}$\pm$9.6 \\
        H-c & 2.1 & 2.1 & 0.4 & 0.6 & 3.0 & 0.3 & 8.9 & 33.8 & \textbf{44.6} & 43.$1\pm$7.6 \\
        D-c & 0.4 & 1.6 & 0.01 & 0.0 & 16.3 & 0.0 & 19.8 & \textbf{23.6} & \textbf{23.6} & 21.5$\pm$2.3\\
        \midrule
        Avg. & 15.6 & 19.7 & 19.8 & 17.5 & 32.2 & 19.4 & 38.9 & 52.5 & 50.1 & \textbf{53.1} \\
        \bottomrule
    \end{tabular} 
    \newline
    \caption{Performance of SlimDT and state-of-the-art baselines on MuJoCo and Adroit tasks. For SlimDT, results are reported as the mean and standard error of normalized returns over $5\times100$ rollouts ($5$ independently training runs using different seeds with $100$ sampled trajectories), generally showing low variance. Hc/Hp/Wk corresponds to the Halfcheetah, Hopper, and Walker tasks, while me/m/mr stands for the medium-expert, medium, and medium-replay datasets. P/H/D corresponds to the Pen, Hammer, and Door tasks, while h/c stands for the human and cloned datasets.}
    \label{tab:mainexp}
\end{table}
As shown in Table \ref{tab:mainexp}, across all tasks in Adroit, SlimDT outperforms DT on every task and is comparable to state-of-the-art offline reinforcement learning algorithms.
To ensure the fairness of comparison, we normalize the scores following the criteria set by D4RL \cite{fu2020d4rl}, where a full score of $100$ represents expert-level performance. The experiment adopts $5$ random seeds, with $100$ trajectories sampled for each set, and the average score is taken as the final result.
\subsection{Ablation Study}

\begin{table}[h]
    \centering
    \begin{tabular}{lrrrrrrrr}
        \toprule
        \textbf{Adroit} & \textbf{Q:N-Pr} & \textbf{Q:N-Po} & \textbf{Q:C-Pr} & \textbf{Q:C-Po} & \textbf{K:N-Pr} & \textbf{K:N-Po} & \textbf{K:C-Pr} & \textbf{K:C-Po} \\
        \midrule
        P-h & 84.7 & 78.9 & 89.3 & \textbf{94.4} & 76.6 & 75.0 & 79.9 & 85.5\\
        H-h & 22.9 & 30.6 & 25.7 & 29.7& \textbf{39.1} & 35.0 & 23.4 & 24.3\\
        D-h & 18.1 & \textbf{21.3} & 19.0 & 19.4 & 19.4 & 19.4 & 18.4 & 19.3 \\
        P-c & \textbf{99.3} & 85.7 & 88.4 & 89.7 & 90.0 & 87.2 & 92.5 & 81.9 \\
        H-c & 19.8 & 16.5 & 16.5 & 0.0 & \textbf{43.1} & 8.5 & 11.0 & 3.4 \\
        D-c & 14.7 & 0.9 & 13.7 & 0.9 & \textbf{21.5} & 7.7 & 11.7 & 6.6\\
        \midrule
        Avg. & 43.3 & 39.0 & 42.1 & 39.0 & \textbf{48.3} & 38.8 & 39.5 & 36.8\\
        \bottomrule
    \end{tabular} 
    \newline
    \caption{Performance comparison of SlimDT with cross-attention on the Adroit benchmark. Pr/Po stand for SlimDT-pre-cond and SlimDT-post-cond respectively; Q/K denote $\text{CrossAttn}_{Q\rightarrow KV}$ and $\text{CrossAttn}_{K\rightarrow QV}$ respectively; N/C represent without causal mask and with causal mask respectively.}
    \label{tab:ablationexp}
\end{table}

\subsubsection{Diffenent Variants of Cross-Attention}
We have experimented with all plausible cross-attention variants on the Adroit task, including $\text{CrossAttn}_{Q\rightarrow KV}$ and $\text{CrossAttn}_{K\rightarrow QV}$ mentioned in Sec. \ref{subsubsec:crossattn}, as well as $\text{CrossAttn}_{KV\rightarrow Q}$, $\text{CrossAttn}_{QV\rightarrow K}$.
Experiments show that only $\text{CrossAttn}_{Q\rightarrow KV}$ and $\text{CrossAttn}_{K\rightarrow QV}$ can achieve satisfactory results, and the latter usually works better.
$\text{CrossAttn}_{KV\rightarrow Q}$, $\text{CrossAttn}_{QV\rightarrow K}$ The two cross-attentions did not converge, so their results are not shown. Possible reasons for the failure are discussed in Sec. \ref{subsubsec:crossattn}.

Besides trying different types of cross-attention, we also experimented with adding causal masks, which enables the target sequence to be modulated only by RTG at preceding time steps. We found that adding causal masks can sometimes improve model performance on high-quality datasets (human). This may be because high-quality datasets impose relatively lower requirements on the expressive capability of the network, making the same architecture more prone to over-parameterization. The introduction of causal masks acts as a parameter regularization mechanism and alleviates the overfitting problem.
However, for low-quality cloned data, adding masks usually leads to performance degradation.

Based on the average performance shown in Tab. \ref{tab:ablationexp}, the pre-cond variant of $\text{CrossAttn}_{K\rightarrow QV}$ without causal mask is ptimal in most cases, exhibiting better generalizability.
\subsubsection{Concatentation and Adaptive Layer Normalization}
We attempted to adopt adaLN as a condition injector on the Adroit benchmark. Although adaLN achieves excellent performance on MuJoCo tasks, it fails to work effectively on Adroit. We attribute this discrepancy to the inherent characteristics of the tasks: the reward signals in MuJoCo are relatively smooth, and relying solely on state-action sequences is sufficient to capture all information required for decision-making. In contrast, tasks in the Adroit suite exhibit highly volatile rewards, where minor changes in state can lead to dramatic reward jumps. As a result, it is difficult to learn an accurate environmental dynamics model using only state-action information. Therefore, effectively leveraging the temporal information embedded in the RTG sequence becomes the key to achieving satisfactory performance. Even when adaLN is used to modulate the state sequence before feeding it into the Transformer, the model's ability to capture critical event information from the RTG sequence within the modulated state sequence remains far inferior to the more straightforward approach of injecting conditional information via cross-attention in a sequence-to-sequence manner.

The failure of adaLN on the Adroit tasks prompted us to explore concatenation-based conditional injection. We first tried directly concatenating the RTG with the state without any mapping, but this approach failed. We then linearly projected the RTG into vectors of varying dimensions, concatenated them with the state, and fed the combined representation into a Transformer for temporal modeling. As shown in Tab. \ref{tab:rtgdim}, this method achieved reasonably good performance, though still inferior to cross-attention.

We observed that better performance is achieved when the RTG is mapped to approximately half of the Transformer's hidden dimension before concatenation with the state. Although this approach does not outperform cross-attention, its performance is close to or better than that of DT. Moreover, the concatenation method is simpler and more efficient in inference.
\begin{table}[h]
    \centering
    \begin{tabular}{lrrrrrrr}
        \toprule 
        Dim & P-h & H-h & D-h & P-c & H-c & D-c & Avg. \\
        \midrule
        32 & 82.7 & 27.1 & 17.1 & 93.8 & 25.9 & 16.7 & 43.8 \\
        64 & \textbf{85.0} & \textbf{28.0} & \textbf{17.2} & \textbf{96.9} & \textbf{27.6} & 18.9 & \textbf{45.6} \\
        128 & 76.0 & 25.0 & 15.8 & 95.6 & 23.0 & 16.2 & 42.0 \\
        \bottomrule
    \end{tabular} 
    \newline
    \caption{SlimDT-pre-cond Performance on Adroit with condition injector: expanding RTG into different dimensions and concatenates it with states.}
    \label{tab:rtgdim}
\end{table}
\subsubsection{Pre-Condition vs. Post-Condition}
We investigate two ways of fusing conditional information, pre-cond and post-cond, on the relatively simple MuJoCo tasks and observe that post-cond achieves significantly better performance. However, as shown in Tab. \ref{tab:ablationexp}, when the dataset is the lower-quality cloned variant, pre-cond obtains a higher chance of superior performance.
In pre-cond, the condition information is further encoded by the Transformer after fusion, which allows more complete extraction of the temporal information contained in the RTG. This property benefits learning when the task is relatively complex or the data quality is low, but becomes a burden when the task is simpler or the data quality is high.
These results inform our choice of condition injection paradigm: in general, pre-cond yields competitive performance, while post-cond may be more effective on simpler tasks or datasets of higher quality.

We also experimented with combining both pre-cond and post-cond injection (based on cross-attention) on the Adroit benchmark. It was observed that this approach tends to yield performance that lies between that of pre-cond injection and post-cond injection individually, without significantly improving the overall performance. Therefore, this scheme is not the preferred choice in practice.

\section{Conclusion and Future Work}
\label{sec:concl}
We presented SlimDT, a simple yet effective modification to the DT that eliminates the Return-to-Go token from the autoregressive sequence and injects its information into state representations via a condition injector. By reducing the sequence length by one third, SlimDT lowers the quadratic inference cost from $9k^2$ to $4k^2$ ($5k^2$ if one cross-attention module is used) while maintaining or improving task performance. Empirical evaluations on D4RL benchmarks: including MuJoCo locomotion and Adroit dexterous manipulation, show that SlimDT consistently outperforms standard DT and achieves competitive results against state-of-the-art offline RL methods. Ablation studies further reveal that the optimal condition injection paradigm (pre-cond vs. post-cond) and injector design (concatenation, adaLN, or cross-attention) depend on task complexity and data quality, offering practical guidelines for practitioners.

While SlimDT achieves strong results across most benchmarks, we note some limitations that point to promising future directions. First, on Adroit tasks with highly volatile rewards, adaLN-based conditioning underperforms compared to cross-attention or concatenation; we suspect this is because adaLN’s point-wise modulation struggles to capture abrupt reward changes that span multiple time steps. A natural extension is to design a hybrid injector that adaptively selects between point-wise and sequence-wise fusion based on reward smoothness. Second, the optimal choice among pre-cond/post-cond and different cross-attention variants still requires task-specific tuning; developing a lightweight heuristic or learned meta-controller to automate this selection remains an open challenge. Third, the proposed conditioning scheme has potential generalizability. Although other sequence models may not suffer from the quadratic inference complexity of Transformers, shortening the sequence length remains beneficial. Nevertheless, further study is needed regarding the adaptation of this scheme to other sequence models.
We believe these limitations are not fundamental and can be addressed in future work without altering the core idea of decoupling sparse conditioning signals from the autoregressive chain.
\small

\bibliographystyle{unsrtnat} 
\bibliography{refs} 

\newpage


\appendix
\newpage
\section{Hyperparameters}

\begin{table}[h]
    \centering
    \begin{tabular}{lr}
        \toprule
        Hyperparameters & Value \\
        \midrule 
        Number of attention layers & $3$ \\
        Attention heads & $1$ \\
        Context length $k$ & $20$ \\
        Batch size & $128$ \\
        Learning rate & $10^{-4}$ \\
        Embedding dimension & $128$ \\
        Dropout rate & $0.1$ \\
        Activation & $\text{ReLU}$ \\
        Weight decay & $10^{-4}$ \\
        Grad norm clip & $0.25$ \\
        LR schedule & Linear warmup for $10^4$ steps \\
        \midrule
        Tasks & RTG Value \\
        \midrule
        Hopper & 7200 \\
        HalfCheetah & 12000 \\
        Walker2d & 5000 \\
        Pen & 10000 \\
        Hammer & 30000 \\
        Door & 6000 \\
    \bottomrule
    \end{tabular}
    \caption{Hyperparameters for both DT and SlimDT.}
    \label{tab:hyperparams}
\end{table}

\section{Experiments Settings}

\begin{table}[h]
    \centering
    \begin{tabular}{lr}
        \toprule
        Setting & Value \\
        \midrule
        CPU & Intel(R) Xeon(R) Gold 6348 CPU @ 2.60GHz; 28 Cores, 112 Threads. \\
        GPU & NVIDIA GeForce RTX 3080, Driver Version: 550.127.05, CUDA Version: 12.4; 8 GPUs. \\
        Memory & 512GB. \\
        OS & 24.04.1 LTS (Noble Numbat). \\
        \midrule
        Total experiment time & 23 days \\
    \bottomrule
    \end{tabular}
    \caption{Computing Resource}
    \label{tab:os}
\end{table}

\end{document}